\documentclass{llncs}
\usepackage{times}
\usepackage{helvet}
\usepackage{courier}
\usepackage{fixbib}
\usepackage{pifont}
\usepackage[vlined,boxed,linesnumbered]{algorithm2e}
\usepackage{graphicx}
\usepackage{latexsym}
\usepackage{amssymb,amsmath}
\usepackage{framed}
\usepackage{tikz}
\usetikzlibrary{arrows,patterns,plotmarks,backgrounds}
 \usepackage{listings} 
\usepackage{subfigure}
\newcommand{\CUM}{\scalebox{.9}[0.95]{\sl Cumulative}}
\newcommand{\RCUM}{{\sl FlexC}}

\newcommand{\fgeMin} {{\sl genEvent\_min}} 
\newcommand{\fgeMax}{{\sl genEvent\_max}} 
\newcommand{\fpeMin} {{\sl processEvent\_min}} 
\newcommand{\fpeMax}{{\sl processEvent\_max}} 
\newcommand{\ffilterMin}{{\sl filter\_min}}
\newcommand{\ffilterMax}{{\sl filter\_max}} 

\newcommand{\SCP}{{\small $SCP$}} 
\newcommand{\ECP}{{\small $ECP$}} 
\newcommand{\PR}{{\small $PR$}} 
\newcommand{\EKCP}{{\small $EKCP$}} 
\newcommand{\KCPI}[1]{KCP_{#1}}

\newcommand{\hEvent} {\mathcal{Q}} 
\newcommand{\hCheck} {h_{check}}

\begin{document}

\title{Dynamic Sweep Filtering Algorithm for \RCUM}
\author{Alban Derrien$^1$, Thierry Petit$^{1,2}$ and St{\'e}phane Zampelli
\institute{
{\bf TR-Mines Nantes: 14/1/INFO} \\
$^1$ TASC (Mines Nantes, LINA, CNRS, INRIA), France \\
$^2$ School of Business, Worcester Polytechnic Institute, USA \\
\{alban.derrien, thierry.petit\}@mines-nantes.fr \\ szampelli@gmail.com \\ tpetit@WPI.edu
}
}
\maketitle
\begin{abstract}
We investigate cumulative scheduling in uncertain environments, using constraint programming. 
We detail in this paper the dynamic sweep filtering algorithm of the \RCUM~global constraint. 
\end{abstract}
\section{Introduction}
When a solution is executed in a real-world environment, 
activities may take longer to execute than expected. 
In many practical cases, solutions cannot be re-computed at anytime when disruptions 
occur. For instance, in Crane Assignment~\cite{DBLP:conf/cp/ZampelliVSDR13}, planners need a fixed schedule which guarantees that
the vessel processing will be completed ahead of schedule. 
The solution should meet the deadline while being able to absorb activity delays during its execution. We wish a tradeoff between robustness and  performance. 

In a Cumulative Scheduling Problem (CuSP), each activity $a \in \mathcal{A}$ has a starting time variable $s_a$ and an ending time variable $e_a$. Its duration $p_a$ (processing time) and resource consumption $h_a$ are usually strictly positive 
integers. We use the notation $a = \langle s_a, p_a, e_a, h_a \rangle$.
Given an integer capacity $C$, a solution to a CuSP satisfies the following constraints:
\begin{eqnarray*}
\forall a \in \mathcal{A}, s_a + p_a = e_a~~~\wedge~~~ 
\forall t \in \mathbb{N}, (\sum_{t \in [s_a, e_a[,a \in \mathcal{A}} h_a) \leq C.
\end{eqnarray*}

In Constraint Programming,  
the \CUM($\mathcal{A},C$)~constraint~\cite{belcon94} represents a CuSP. 
A usual objective is to minimize the \emph{makespan}, i.e., the latest end among all activities. 

In this paper, we investigate the filtering algorithm of the \RCUM~constraint~\cite{derpetzam14}. \RCUM~expresses cumulative problems that should integrate a given level of robustness. 
Thus, \RCUM~represents a new problem, derived from the CuSP. In order to define this problem 
We use the following notation  for $i$-order maximum heights of activities: Given  $\mathcal{A}^\downarrow$ the collection of activities in a set $\mathcal{A}$ sorted by decreasing heights, 
$\mathit{max}_{a \in \mathcal{A}}^i(h_a)$ is the height of the $i^{th}$ activity in $\mathcal{A}^\downarrow$. 

\begin{definition}[RCuSP$^r$] \label{def:r1}
Given a set of activities $\mathcal{A}$, let $\mathcal{K}$ be a set of positive integers slacks associated with activities,
such that to each $a \in \mathcal{A}$ corresponds $k_a \in \mathcal{K}$. Let $r$ be an integer, $r\geq 1$.
A solution to a \emph{RCuSP}$^r$ satisfies the following constraints: 
\vspace{-3mm}
\begin{eqnarray*}
 \forall a \in \mathcal{A}, s_a + p_a = e_a ~~~\wedge~~~\forall t \in \mathbb{N}, \sum \limits_{  \underset{t \in [s_a, e_a[}{a \in \mathcal{A},} } h_a
     + \sum \limits_{i=1}^{i=r}
      \mathop{\mathit{max}^i}\limits_{a \in \{b\in\mathcal{A}, t \in [e_b,e_b+k_b[\}} (h_a)
         \leq C
\end{eqnarray*}
\vspace{-6mm}
\end{definition}

We focus on the problem RCuSP (RCuSP$^r$ with $r=1$). RCuSP is the problem encoded by the constraint \RCUM($\mathcal{A}, C, \mathcal{K}$).  
\section{Dynamic Sweep for \RCUM}
This section presents a Time-Table filtering algorithm for \RCUM, stem from the Dynamic Sweep algorithm for CuSP~\cite{letbelcar12}. 
This algorithm prunes starting time variables and ending time variables of activities. It reduces the bounds of domains and does not directly depends on the time unit.
Given a variable $x$, $\underline{x}$ (resp. $\overline{x}$) denotes the minimum value (resp. the maximum value) in its domain.
\subsection{Failure and Fix-Point Conditions}
In this section we recall the conditions that are exploited by the algorithm. Proofs and explanations can be found in~\cite{derpetzam14}. 
\begin{definition}[$\mathcal{K}$-compulsory part~{~\cite{derpetzam14}}] \label{def:KCP} 
Let $a \in \mathcal{A}$ be an activity and $k_a \in \mathcal{K}$.
The $\mathcal{K}$\emph{-compulsory part} of $a$, denoted $\KCPI{a}$, is the interval $[\max(\overline{s_a},\underline{e_a}), \underline{e_a}+k_a]$.  
\end{definition}

The Time-Table failure condition of \RCUM~integrates in the profile, at any time $t$, the maximum 
height among activities having a $\mathcal{K}$-compulsory part intersecting $t$. 
{\begin{sloppypar}
\begin{proposition}[Time-Table failure check for \RCUM~{~\cite{derpetzam14}}]~\\\label{prop:chk}
If $\exists$$t$$\in$$\mathbb{N},   
 (\sum_{ {a \in \mathcal{A}}, {t \in [\overline{s_a}, \underline{e_a}[} } h_a)$$+$$(\max_{ { a \in \mathcal{A}, {t \in \KCPI{a}}}} h_a)$$>$$C$ then \RCUM($\mathcal{A}, C, \mathcal{K}$)~fails.    
\end{proposition}
\end{sloppypar}
}
\begin{definition}

Given a scheduling constraint, a propagator is \emph{Time-Table} if $\forall a \in \mathcal{A}$, fixing $s_a$ at time $\underline{s_a}$ (respectively, $e_a$ at time $\overline{e_a}$) does not lead to 
 a contradiction if we apply the Time-Table Failure check of the constraint. 
\end{definition}

The following property holds when  Letort et al.'s \emph{sweep\_min} algorithm reaches its fixpoint (Property 1 in~\cite{letbelcar12}) on lower-bounds of start variables. 
\begin{property}[\CUM~(lower bounds) \emph{~\cite{letbelcar12}}]\label{prop:sweep}
Given \CUM($\mathcal{A},C$), the time-table propagator ensures that 
$\forall b \in \mathcal{A}$:  $$\forall t \in [\underline{s_b}, \underline{e_b}[, h_b + \sum_{  {a \in \mathcal{A} \setminus \{ b \}}, {t \in [\overline{s_{a}}, \underline{e_{a}}[} } h_{a} \leq C$$ 
\end{property}

The complete Time-Table fixpoint conditions for \RCUM are the following. Any activity which would lead to a Time-Table fail if fixed at its earliest (resp. latest) date violates one of the conditions, and reciprocally.  

\begin{property}[\RCUM~(lower bounds)~\emph{\cite{derpetzam14}}]\label{prop:sweeprcum}
Given \RCUM($\mathcal{A},C,\mathcal{K}$), the propagator should ensure $\forall b \in \mathcal{A}$:
\begin{equation*} 
 {
   \begin{array}{ll}
    \forall t \in [\underline{s_b}, \underline{e_b}[,
 (h_b +\sum \limits_{  \underset{t \in [\overline{s_{a}}, \underline{e_{a}}[} {a \in \mathcal{A} \setminus \{ b \},}} h_{a}) 
 + 
 (\max \limits_{ \underset{t \in \KCPI{a}} {a \in \mathcal{A},}} h_{a}) 
~\leq~C  &~~~(1) \\
\wedge
 \forall t \in [\underline{e_b}, \underline{e_b}+k_b[,
 (\sum \limits_{  \underset{t \in [\overline{s_{a}}, \underline{e_{a}}[} {a \in \mathcal{A}, }} h_{a}) 
 +
 h_b 
 ~\leq~ C &~~~(2)  
   \end{array}
}
\end{equation*}
\end{property}
\begin{property}[\RCUM~(upper bounds)~\emph{\cite{derpetzam14}}]\label{prop:sweeprcum-max}
Given \RCUM($\mathcal{A},C,\mathcal{K}$), the propagator should ensure the same 
conditions as Property~\ref{prop:sweeprcum} with intervals $[\overline{s_b}, \overline{e_b}[$  
(condition (1)) and $[\overline{e_b}, \overline{e_b}+k_b[$ (condition (2)).
\end{property}
\subsection{Filtering Algorithm}
This section details the modifications that are required to adapt Letort et al. algorithm for \CUM~\cite{letbelcar12} to the case of \RCUM. 
This algorithm is in two steps: Filtering of lower bounds of starting time variables ({\sl Sweep\_min}) and 
upper bounds of ending-time variables ({\sl Sweep\_max}). 
\subsubsection{Background: Sweep-min for Cumulative}
The principle is to move a sweep line from the earliest starting time to the end of the
schedule. Two consecutive steps correspond to two consecutive changes in the profile of compulsory parts. 
The data exploited at the current position $\delta$ of the sweep line is the height of the profile $ph_\delta$. 
At  $\delta$, the algorithm uses events\footnote{A triplet $<$type$\in$$\{${\small $SCP,ECP,PR$}$\}$, activity, date$>$.} stored in a queue $\mathcal{Q}$. 
The event types are: \SCP~(start of a compulsory part, at date $\overline{s_a}$), \ECP~(end of a compulsory part, at date $\underline{e_a}$), and \PR, 
which indicates that an activity is candidate for filtering, stored in a data structure $\hCheck$.

\paragraph{Static version. }
In the static {\sl Sweep\_min} algorithm, all events are computed from scratch and added to $\mathcal{Q}$ before the sweep, by a {\sl generateEvents} procedure. They are progressively 
removed from $\mathcal Q$ and treated, while the sweep line moves from the earliest event to the latest one (on the right) in a {\sl sweepMin} procedure.  
All events at date $\delta$ are  processed to compute the profile height $ph_\delta$, which is constant up to the next event date $\delta'$.
From Property~\ref{prop:sweep}, an activity $a \in \hCheck$ is pruned if scheduling that one at its earliest date $\underline{s_a}$ leads to $h_a + ph_\delta > C$.
In this case, $\underline{s_a}$ is adjusted to $\delta'$. The whole process {\sl generateEvents} $+$ {\sl sweepMin}  is repeated while  at least one adjustment has been performed. 

\paragraph{Dynamic version. }
We describe the version of Letort's PhD disseration~\cite{let13}. 
The idea is that it is possible to update compulsory parts on the fly, without creating any compulsory part before (on the left of) the current position $\delta$. 
The events queue $\mathcal Q$ is thus dynamic. 
Given an activity $a$, as the technique only adjusts lower bounds of variables, $\overline{s_a}$ does not change. 
Therefore, \SCP~events are generated for all activities in the \fgeMin~procedure, even if they do not have initially a 
compulsory part. In our implementation, we state definitively the existence of a compulsory part for an activity $a$ when the \SCP~event is handled: 
\begin{proposition}\label{prop:decisionsafter}
In the \fpeMin~procedure, at time $\overline{s_a}$  all the decisions with respect to activity $a$ can be taken. 
\end{proposition}
When a \SCP~event is handled at date $\delta$, if the activity $a$ has a compulsory part, then the corresponding \ECP~event is dynamically created and the height of the activity is added to $ph_\delta$. 
The pruning rule is the same as in the static version.
\footnote{In~\cite{letbelcar12,let13} candidates for pruning are separated in two sets ($h_\mathit{check}$ and $h_\mathit{conflict}$). 
In the new version of Letort's PhD dissertation, this separation is actually not mandatory but optimizes the code. 
To simplify the presentation, we use only one set.} 
This algorithm is able to reach its fix point in a single step. 
The general scheme is described in Algorithm~\ref{algo:dyna}.

\begin{algorithm}
{
	\fgeMin()\;
	\While{$\mathcal{Q}$ is not empty}{
	    \fpeMin()\;
	    \ffilterMin()\;
	}
	\caption{{\sl Sweep\_min}().}
	\label{algo:dyna}
}
\end{algorithm}

\subsubsection{Modified Sweep\_min for \RCUM.}
We integrate into the reasoning $\mathcal{K}$-compulsory parts. 
To reach the fix point in a single step, we must not create a $\mathcal{K}$-compulsory part before (on the left of) the sweep line. If we violate 
this rule, some data previously computed should not remain valid at the current position $\delta$ of the sweep line. 
The $\mathcal{K}$-compulsory part of an activity $a$ is $[\max(\overline{s_a},\underline{e_a}), \underline{e_a}+k_a]$. 
From Proposition~\ref{prop:decisionsafter}, at time $\overline{s_a}$ this interval is known.
Since this interval is after $\overline{s_a}$, no $\mathcal{K}$-compulsory part is created on the left of the sweep line. 

In the following, $\mathcal{L}$ denotes a heap of activities for which $\delta$ is in the $\mathcal{K}$-compulsory part, ordered by decreasing heights. 
Adding a new element is usual. Conversely, removing is done lazily when we get the head (an activity with the maximum height). We first describe the corresponding function, 
$\max(\mathcal{L})$.
\begin{algorithm}
{
	\lIf{$\mathcal{L}$.isEmpty()}{
	\Return 0\;
	}
	$\mathit{peeka} \leftarrow \mathcal{L}$.peek()\;
	\While{$\delta \leq \overline{s_{\mathit{peeka}}} \vee \delta \leq \underline{e_{\mathit{peeka}}} $}{
		$\mathcal{L}$.removePeek()\; 
		\lIf{$\mathcal{L}$.isEmpty()}{
		\Return 0\;
		}
		$\mathit{peeka} \leftarrow \mathcal{L}$.peek()\;
	}
	\Return $h_{\mathit{peeka}}$\;
	\caption{$\max(\mathcal{L})$: Integer}
	\label{algo:maxl}
}
\end{algorithm}

We use a new class of events, \EKCP, which indicates the end of a $\mathcal{K}$-compulsory part. \\\\
\emph{$(1)$ \fgeMin~procedure.} \\ 

From the set of activities~$\mathcal{A}$, it generates \SCP~events at time $\overline{s_a}$ and \PR~events, at time $\underline{s_a}$, for activities which are candidate for pruning.
{
\begin{algorithm}
{
	\ForEach{$a \in \{\mathcal{A}\}$}{
		$\mathcal{Q}  \leftarrow \mathcal{Q} \cup \{<${$SCP$}$, a, \overline{s_a}\!>\}$\;
		\lIf{$\underline{s_a} \neq \overline{s_a}$}{
			$\mathcal{Q}  \leftarrow \mathcal{Q} \cup \{<${$PR$}$, a, \underline{s_a}\!>\}$\;
		}
	}
	\caption{\fgeMin()}
	\label{algo:genmin}
}
\end{algorithm}
}

\noindent\emph{$(2)$ Handling the start of a $\mathcal{K}$-compulsory part, $max(\overline{s_a},\underline{e_a})$.}\\

If activity $a$ has a compulsory part, its $\mathcal{K}$-compulsory part starts at the end of this compulsory part $\underline{e_a}$. 
Therefore, when the \ECP~event of activity $a$ is handled, we add $a$ in $\mathcal{L}$ and 
we add a new event \EKCP~in $\mathcal{Q}$ at date $\underline{e_a}+k_a$. 

Otherwise,  $a$ may have a $\mathcal{K}$-compulsory part, if $\underline{e_a}+k_a>\overline{s_a}$. 
This situation is detected when the \SCP~event of $a$ is handled (at time $\overline{s_a}$). 
In this case, this compulsory part starts at $\delta$ and we add $a$ in $\mathcal{L}$. 
We add a new event \EKCP~in $\mathcal{Q}$ at date $\underline{e_a}+k_a$. \\\\
\emph{$(3)$ Handling the end of a $\mathcal{K}$-compulsory part, $\underline{e_a}+k_a$.} \\

Nothing to do as removing activity $a$ from $\mathcal{L}$ is done lazily. \\
\begin{algorithm}
{
\linespread{0,7}
$(\delta,\xi) \leftarrow$ extract and record in a set $\xi$ all events in $\mathcal{Q}$ related to the minimal date $\delta$

	\ForEach{ events of type $<\!$ {$SCP$}$, a, \overline{s_a}\!>$ in $\xi$}{
		\uIf{$\delta < \underline{e_a}$}{
			$ph_{\delta}~ +\!\!= h_a$\;
			$\hEvent \leftarrow \hEvent \cup \{<\!$ {$ECP$}$, a, \underline{e_a}\!>\} $\;
		}
		\ElseIf{ $\delta < \underline{e_a} + k_a$ }{
			$\mathcal{L} \leftarrow \mathcal{L} \cup \{a\}$\;
			$\hEvent \leftarrow \hEvent \cup \{<\!$ {$EKCP$}$, a, \underline{e_a} \!+\! k_a\!>\}$\;
		}		
	}
	\ForEach{ events of type $<\!$ {$ECP$}$, a, \underline{e_a}\!>$ in $\xi$}{		
		$ph_{\delta}~ -\!\!= h_a$\;
		$\mathcal{L} \leftarrow \mathcal{L} \cup \{a\}$\;
		$\hEvent \leftarrow \hEvent \cup \{<\!$ {$EKCP$}$, a, \underline{e_a} \!+\! k_a\!>\}$\;		
	}
		\lForEach{ events of type $<\!$ {$PR$} $, a, \underline{s_a}\!>$ in $\xi$}{					
		$\hCheck \leftarrow \hCheck \cup \{a\}$\;		
	}
	\caption{\fpeMin()}
	\label{algo:peMin}
}
\vspace{0.1cm}
\end{algorithm}

\noindent\emph{$(4)$ Fitering.}\\

{
\begin{algorithm}
{
\ForEach{$a$ in $\hCheck$}{
\lIf{$\underline{e_a}+k_a = \delta$}{
	$\hCheck \leftarrow \hCheck \setminus \{ a \}$\;
}
\Else{
\lIf{$( \underline{e_a} \le \delta) \wedge (ph_\delta + h_a > C)$}{
	$\underline{s_a} \leftarrow \delta'$\;
}
\lIf{$(\delta' \le \underline{e_a}) \wedge (ph_\delta + h_a + \max(\mathcal{L}) > C) $}{
	$\underline{s_a}\leftarrow \delta'$ \;
}
}
}
\caption{ \ffilterMin() }\label{alg:filtermin}
}
\end{algorithm}
}

\emph{Time complexity.} Recall we use a heap as a data structure $\mathcal{L}$, from which activities are added and removed only once per sweep. Getting the head can be done in constant time.
The filtering procedure has the same time complexity as in the case of 
\CUM. Therefore, {\sl Sweep\_min} for \RCUM~is in $O(n^2)$ time, as for \CUM~\cite[p. 55]{let13}. 

\subsubsection{Modified Sweep\_max for \RCUM.}

Conversely to the case of \CUM, the filtering of \RCUM~is not symmetrical. 
We present the solution we have designed to obtain a dynamic filtering of upper bounds regarding ending time variables. 
 {\sl Sweep\_max}  
sweeps from the right to the left. \\\\
\emph{$(1)$ \fgeMax~procedure.} \\ 

From~$\mathcal{A}$, it generates \EKCP~events at time $\underline{e_a}\!+\!k_a$ ,~\ECP~events at time $\underline{e_a}\!$~and~\PR~events, at time $\overline{e_a}\!+\!k_a$, for activities which are candidate for pruning.\\\\
{
\begin{algorithm}
{
	\ForEach{$a \in \{\mathcal{A}\}$}{
		$\mathcal{Q}  \leftarrow \mathcal{Q} \cup \{<${$EKCP$}$, a, \underline{e_a}\!+\!k_a\!>\}$\;
		$\mathcal{Q}  \leftarrow \mathcal{Q} \cup \{<${$ECP$}$, a, \underline{e_a}\!>\}$\;

		\lIf{$\underline{s_a} \neq \overline{s_a}$}{
			$\mathcal{Q}  \leftarrow \mathcal{Q} \cup \{<${$PR$}$, a, \overline{e_a}\!+\!k_a\!>\}$\;
		}
	}
	\caption{\fgeMax()}
	\label{algo:genmax}
}
\end{algorithm}
}

\noindent\emph{$(2)$ Handling the end of a $\mathcal{K}$-compulsory part, $\underline{e_a}+k_a$.}\\

When at position $\delta$ {\sl \fpeMax} manages an \EKCP~event, it is necessary to verify whether a $\mathcal{K}$-compulsory part exists 
for the corresponding activity $a$, or not. 

If $\overline{s_a} \geq \delta$  a valid support for $\overline{e_a}$ has been found, and this activity does not have a $\mathcal{K}$-compulsory part.
Nothing has to be done. 

Otherwise, it exists a $\mathcal{K}$-compulsory part starting at $t = \max(\overline{s_a}, \underline{e_a})$. We create a new event $SKCP$ at this date $t$ and we 
add $a$ to $\mathcal{L}$. \\\\
\emph{$(3)$ Handling the start of a $\mathcal{K}$-compulsory part, $max(\overline{s_a},\underline{e_a})$.}\\\\
When the \fpeMax~procedure manages a $SKCP$ event, we verify that no filtering has been made on the activity $a$ since this event was created. 

If $a$ was filtered, it is necessary to create a new $SKCP$ event for $a$ at $t = \max(\overline{s_a}, \underline{e_a})$. \\

\begin{algorithm}
{
$(\delta,\xi) \leftarrow$ extract and record in $\xi$ all event in $\mathcal{Q}$ related to the maximal date $\delta$

	\ForEach{ events of type $<\!$ {$ECP$}$, a, \underline{e_a}\!>$ in $\xi$}{
		\If{$ \overline{s_a}<\delta $}{
			$ph_{\delta}~ +\!\!= h_a$\;
			$\hEvent \leftarrow \hEvent \cup \{<${$SCP$}$, a, \overline{s_a}\!>\}$\;
		}	
	}
	\lForEach{ events of type $<\!$ {$SCP$}$, a, \overline{s_a}>$ in $\xi$}{		
		$ph_{\delta}~ -\!\!= h_a$\;
	}
	\ForEach{ events of type $<\!$ {$EKCP$}$, a, \underline{e_a}\!+\!k_a\!>$ in $\xi$}{
		\If{$max(\overline{s_a},\underline{e_a})<\delta$}{
			$\mathcal{L} \leftarrow \mathcal{L}  \cup \{a\}$\;
			$\hEvent \leftarrow \hEvent \cup \{<\!$ {$SKCP$}$, a, \max(\overline{s_a},\underline{e_a})\!>\}$\;
		}	
	}
	\ForEach{ events of type $<\!$ {$SKCP$}$, a, \overline{s_a}\!>$ in $\xi$}{
		\lIf{$max(\overline{s_a},\underline{e_a})<\delta$}{
			$\hEvent \leftarrow \hEvent \cup \{<\!$ {$SKCP$}$, a, \max(\overline{s_a},\underline{e_a})\!>\}$\;
		}
	}
	\lForEach{ events of type $<\!$ {$PR$}$, a, \overline{e_a} \!+\! k_a\!>$ in $\xi$}{			
		$\hCheck \leftarrow \hCheck \cup \{a\}$\;		
	}
	\caption{\fpeMax()}
	\label{algo:peMax}
}
\end{algorithm}

\noindent\emph{$(4)$ Filtering.} \\

Compared with {\sl Sweep\_min}, 
an important difference is that, for an activity $a$, when the current interval is in the $\mathcal{K}$-compulsory part of $a$ 
then $a$ must not be taken into account in its own pruning condition.
To express this regret mechanism, we 
use a function $\max_a(\mathcal{L})$, which returns the height of either the activity with maximum height if it is not $a$, or  
the second maximum otherwise.\footnote{As $\max(\mathcal{L})$, the second activity must also be checked and removed if needed.}  
{
\begin{algorithm}

\caption{ \ffilterMax()}\label{alg:filtermax}
\ForEach{$a$ in $\hCheck$}{
	\lIf{$\delta = \overline{s_a} $}{
			$\hCheck \leftarrow \hCheck \setminus \{ a \}$\;
	}
	\Else{
		\lIf{$( \overline{e_a} \le \delta' ) \wedge (ph_\delta + h_a > C)$}{
			$\overline{e_a} \leftarrow \delta' - k_a$\;
		}
		\lIf{$(\delta \le \overline{e_a}) \wedge (ph_\delta + h_a + \max_a(\mathcal{L}) > C) $}{
			$\overline{e_a} \leftarrow \delta'$ \;
		}
		}
	}
\end{algorithm}
}
%

\emph{Time complexity.} 
Conversely to {\sl Sweep\_min},  $SKCP$ events in {\sl Sweep\_max}  can be created several times for a given activity $a$. However, the maximum number of generations 
is bounded by a value $X \leq k_a$. Therefore, {\sl Sweep\_max}~for \RCUM~deals with $O(n\times\max_{a \in \mathcal{A}}(k_a))$ events. 
Its time complexity is $O(n^2\times\max_{a \in \mathcal{A}}(k_a)))$. 

\section{Experiments}
As some differences exist with Sweep for \CUM~(no symmetrical algorithms, new events are added), we experimented the limits of our algorithm with respect to problem size. 
We used Choco~\cite{cho13} with OSX 10.8.5, a 2.9 Ghz Intel i7 and 8GB of RAM. 
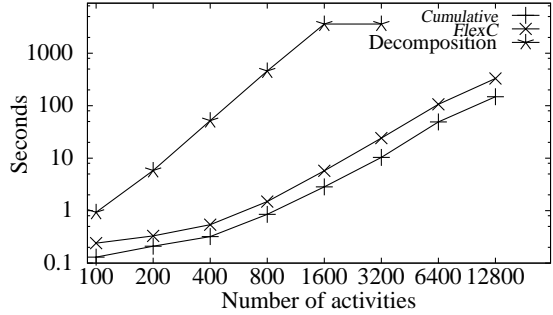
\begin{figure}[!h]
\begin{center}
\begin{tikzpicture}[scale=.4]
\path (0.000,0.000) rectangle (18.000,10.000);
\draw[solid] (1.688,0.985)--(1.868,0.985);
\draw[solid] (17.079,0.985)--(16.899,0.985);
\node[anchor=mid east,yshift=-.12ex] at (1.504,0.985) {\small  0.1};
\draw[solid] (1.688,1.510)--(1.778,1.510);
\draw[solid] (17.079,1.510)--(16.989,1.510);
\draw[solid] (1.688,2.205)--(1.778,2.205);
\draw[solid] (17.079,2.205)--(16.989,2.205);
\draw[solid] (1.688,2.561)--(1.778,2.561);
\draw[solid] (17.079,2.561)--(16.989,2.561);
\draw[solid] (1.688,2.730)--(1.868,2.730);
\draw[solid] (17.079,2.730)--(16.899,2.730);
\node[anchor=mid east,yshift=-.12ex] at (1.504,2.730) {\small  1};
\draw[solid] (1.688,3.256)--(1.778,3.256);
\draw[solid] (17.079,3.256)--(16.989,3.256);
\draw[solid] (1.688,3.950)--(1.778,3.950);
\draw[solid] (17.079,3.950)--(16.989,3.950);
\draw[solid] (1.688,4.306)--(1.778,4.306);
\draw[solid] (17.079,4.306)--(16.989,4.306);
\draw[solid] (1.688,4.475)--(1.868,4.475);
\draw[solid] (17.079,4.475)--(16.899,4.475);
\node[anchor=mid east,yshift=-.12ex] at (1.504,4.475) {\small  10};
\draw[solid] (1.688,5.001)--(1.778,5.001);
\draw[solid] (17.079,5.001)--(16.989,5.001);
\draw[solid] (1.688,5.695)--(1.778,5.695);
\draw[solid] (17.079,5.695)--(16.989,5.695);
\draw[solid] (1.688,6.051)--(1.778,6.051);
\draw[solid] (17.079,6.051)--(16.989,6.051);
\draw[solid] (1.688,6.221)--(1.868,6.221);
\draw[solid] (17.079,6.221)--(16.899,6.221);
\node[anchor=mid east,yshift=-.12ex] at (1.504,6.221) {\small  100};
\draw[solid] (1.688,6.746)--(1.778,6.746);
\draw[solid] (17.079,6.746)--(16.989,6.746);
\draw[solid] (1.688,7.440)--(1.778,7.440);
\draw[solid] (17.079,7.440)--(16.989,7.440);
\draw[solid] (1.688,7.797)--(1.778,7.797);
\draw[solid] (17.079,7.797)--(16.989,7.797);
\draw[solid] (1.688,7.966)--(1.868,7.966);
\draw[solid] (17.079,7.966)--(16.899,7.966);
\node[anchor=mid east,yshift=-.12ex] at (1.504,7.966) {\small  1000};
\draw[solid] (1.688,8.491)--(1.778,8.491);
\draw[solid] (17.079,8.491)--(16.989,8.491);
\draw[solid] (1.688,9.186)--(1.778,9.186);
\draw[solid] (17.079,9.186)--(16.989,9.186);
\draw[solid] (1.688,9.542)--(1.778,9.542);
\draw[solid] (17.079,9.542)--(16.989,9.542);
\draw[solid] (1.976,0.985)--(1.976,1.165);
\draw[solid] (1.976,9.631)--(1.976,9.451);
\node[anchor=mid,yshift=-.12ex] at (1.976,0.50) {\small 100};
\draw[solid] (3.872,0.985)--(3.872,1.165);
\draw[solid] (3.872,9.631)--(3.872,9.451);
\node[anchor=mid,yshift=-.12ex] at (3.872,0.5) {\small 200};
\draw[solid] (5.768,0.985)--(5.768,1.165);
\draw[solid] (5.768,9.631)--(5.768,9.451);
\node[anchor=mid,yshift=-.12ex] at (5.768,0.5) {\small 400};
\draw[solid] (7.664,0.985)--(7.664,1.165);
\draw[solid] (7.664,9.631)--(7.664,9.451);
\node[anchor=mid,yshift=-.12ex] at (7.664,0.5) {\small 800};
\draw[solid] (9.560,0.985)--(9.560,1.165);
\draw[solid] (9.560,9.631)--(9.560,9.451);
\node[anchor=mid,yshift=-.12ex] at (9.560,0.5) {\small 1600};
\draw[solid] (11.456,0.985)--(11.456,1.165);
\draw[solid] (11.456,9.631)--(11.456,9.451);
\node[anchor=mid,yshift=-.12ex] at (11.456,0.5) {\small 3200};
\draw[solid] (13.352,0.985)--(13.352,1.165);
\draw[solid] (13.352,9.631)--(13.352,9.451);
\node[anchor=mid,yshift=-.12ex] at (13.352,0.5) {\small 6400};
\draw[solid] (15.248,0.985)--(15.248,1.165);
\draw[solid] (15.248,9.631)--(15.248,9.451);
\node[anchor=mid,yshift=-.12ex] at (15.248,0.5) {\small 12800};
\draw[solid] (1.688,9.631)--(1.688,0.985)--(17.079,0.985)--(17.079,9.631)--cycle;
\node[anchor=mid,yshift=-.12ex,rotate=-270] at (-0.5,5.308) {Seconds};
\node[anchor=mid,yshift=-.12ex] at (9.383,-0.3) {Number of activities};
\node[anchor=mid east,yshift=-.12ex,font={\fontsize{8pt}{9.6pt}\selectfont}] at (15.611,9.226) {\scriptsize \CUM};
\draw[color=black,solid] (15.795,9.226)--(16.711,9.226);
\draw[color=black,solid] (1.976,1.184)--(3.872,1.547)--(5.768,1.867)--(7.664,2.607)--(9.560,3.521)%
  --(11.456,4.501)--(13.352,5.681)--(15.248,6.519);
\path[color=black,solid] plot[only marks,mark size=8,mark=+] coordinates {(1.976,1.184)};
\path[color=black,solid] plot[only marks,mark size=8,mark=+] coordinates {(3.872,1.547)};
\path[color=black,solid] plot[only marks,mark size=8,mark=+] coordinates {(5.768,1.867)};
\path[color=black,solid] plot[only marks,mark size=8,mark=+] coordinates {(7.664,2.607)};
\path[color=black,solid] plot[only marks,mark size=8,mark=+] coordinates {(9.560,3.521)};
\path[color=black,solid] plot[only marks,mark size=8,mark=+] coordinates {(11.456,4.501)};
\path[color=black,solid] plot[only marks,mark size=8,mark=+] coordinates {(13.352,5.681)};
\path[color=black,solid] plot[only marks,mark size=8,mark=+] coordinates {(15.248,6.519)};
\path[color=black,solid] plot[only marks,mark size=8,mark=+] coordinates {(16.253,9.226)};
\node[anchor=mid east,yshift=-.12ex,font={\fontsize{8pt}{9.6pt}\selectfont}] at (15.611,8.776) {\scriptsize \RCUM};
\draw[color=black,solid] (15.795,8.776)--(16.711,8.776);
\draw[color=black,solid] (1.976,1.649)--(3.872,1.890)--(5.768,2.263)--(7.664,3.032)--(9.560,4.056)%
  --(11.456,5.143)--(13.352,6.267)--(15.248,7.126);
\path[color=black,solid] plot[only marks,mark size=8,mark=x] coordinates {(1.976,1.649)};
\path[color=black,solid] plot[only marks,mark size=8,mark=x] coordinates {(3.872,1.890)};
\path[color=black,solid] plot[only marks,mark size=8,mark=x] coordinates {(5.768,2.263)};
\path[color=black,solid] plot[only marks,mark size=8,mark=x] coordinates {(7.664,3.032)};
\path[color=black,solid] plot[only marks,mark size=8,mark=x] coordinates {(9.560,4.056)};
\path[color=black,solid] plot[only marks,mark size=8,mark=x] coordinates {(11.456,5.143)};
\path[color=black,solid] plot[only marks,mark size=8,mark=x] coordinates {(13.352,6.267)};
\path[color=black,solid] plot[only marks,mark size=8,mark=x] coordinates {(15.248,7.126)};
\path[color=black,solid] plot[only marks,mark size=8,mark=x] coordinates {(16.253,8.776)};
\node[anchor=mid east,yshift=-.12ex,font={\fontsize{8pt}{9.6pt}\selectfont}] at (15.611,8.326) {Decomposition };
\draw[color=black,solid] (15.795,8.326)--(16.711,8.326);
\draw[color=black,solid] (1.976,2.675)--(3.872,4.066)--(5.768,5.724)--(7.664,7.376)--(9.560,8.937)%
  --(11.456,8.937);
\path[color=black,solid] plot[only marks,mark size=8,mark=star] coordinates {(1.976,2.675)};
\path[color=black,solid] plot[only marks,mark size=8,mark=star] coordinates {(3.872,4.066)};
\path[color=black,solid] plot[only marks,mark size=8,mark=star] coordinates {(5.768,5.724)};
\path[color=black,solid] plot[only marks,mark size=8,mark=star] coordinates {(7.664,7.376)};
\path[color=black,solid] plot[only marks,mark size=8,mark=star] coordinates {(9.560,8.937)};
\path[color=black,solid] plot[only marks,mark size=8,mark=star] coordinates {(11.456,8.937)};
\path[color=black,solid] plot[only marks,mark size=8,mark=star] coordinates {(16.253,8.326)};
\draw[solid] (1.688,9.631)--(1.688,0.985)--(17.079,0.985)--(17.079,9.631)--cycle;
\end{tikzpicture}
\end{center}
\vspace{-0.5cm}
\caption{\small Scaling of Dynamic Sweep for \RCUM.}
\label{fig:bench2}
\end{figure}

Following experiments provided in~\cite{let13}, we generated large (simple) random instances with $p_a$ from $5$ to $10$, $h_a$ from $1$ to $5$, $C = 30$. 
Values in $\mathcal{K}$ are not null, with an average equal to $4$. Similar results are obtained with fixed $k_a$. Figure~\ref{fig:bench2} shows that 
our filtering algorithm scales on problems up 12800 activities for a first solution. 
The decomposition reaches the time limit of 1h:00m 
with 1600 activities and leads to a memory crash with 6400 \CUM. 

In a second experiment,  we evaluate the performance of our approach. We find optimal solutions of cumulative problems where the goal 
is to minimize the makespan. To express robustness, we use either \RCUM~or the naive approach 
which consists of augmenting directly the duration $p_a$ of any activity $a \in \mathcal{A}$ by its corresponding value $k_a \in \mathcal{K}$. 
We solve optimally 50 random problems with $10$ activities, all with $p_a$ from $1$ to $9$ and $h$ from $1$ to $5$. The capacity $C$ is fixed to $16$. 
Each problem is solved for all values of $k_a$ from $1$ to $10$ (same $k_a$ for all activities), to show the impact of $k_a$ 
in comparison with the lengths of activities. 

Table~\ref{fig:bench1} shows the minimum, average and maximum deviation between optimal makespans using \RCUM~and the naive approach, 
normalized with the makespan of the original cumulative problem.\footnote{($obj$ of the naive approach $-$ $obj$ of \RCUM) $/$ $obj$ of \CUM.} 
For the naive approach, the makespan is $\max_{a \in \mathcal{A}}(e_a)$. With respect to \RCUM, the makespan 
is the worst case scenario, i.e., $\max_{a \in \mathcal{A}}(e_a+k_a)$. Table~\ref{fig:bench1} also indicates the number of instances 
for which \RCUM~is worse (respectively better) than the naive approach. 
{
\begin{table}[!h]
{
\begin{center}
\begin{tabular}{|l|r|r|r|r|r|r|r|r|r|r|}\hline
$k_a/p_a$			&	0.2		&	0.4		&	0.6		&	0.8		&	1		&	1.2		&	1.4		&	1.6		&	1.8		&	2 \\\hline
Min         			&	0.0		&	0.0		&	5.9		&	13.0		&	21.7		&	27.3		&	29.4		&	29.4		&	29.4		&	29.4\\\hline
Avg  			&	3.9		&	10.5		&	20.0		&	29.1		&	38.3		&	48.8		&	60.4		&	72.5		&	85.6		&	98.0\\\hline
Max  			&	10.0		&	18.2		&	50.0		&	62.5		&	87.5		&	112.5	&	137.5	&	150.0	&	175.0	&	187.5\\\hline
\#Worse	&	0		&	0		&	0		&	0		&	0		&	0		&	0		&	0		&	0		&	0\\\hline
\#Better 	&	31		&	49		&	50		&	50		&	50		&	50		&	50		&	50		&	50		&	50\\\hline
\end{tabular}
\end{center}
}
\caption{Comparison of makespan for optimal solutions of random problems.}\label{fig:bench1}
\end{table}
}

The results show that the ratio of objective values is significantly in favor of the use of \RCUM, even with a small number of activities. 
We observed a similar behavior with different parameters, such as $h$ varying from $1$ to $10$, and capacities $C=9$ and $C=25$. 
We selected in Table~\ref{fig:bench1} the less favorable results for \RCUM, with small activities and problems. Especially, with $h$ varying from $1$ to $10$, the gain with \RCUM~is higher.

\section{Conclusion}
We investigated a new solving technique for the declarative framework presented in~\cite{derpetzam14}.  
We provided a new dynamic sweep filtering algorithm of the \RCUM~global constraint. Our experiments show that this new filtering algorithm is a good candidate for providing first robust 
solutions on large instances of RCuSP. 
\bibliographystyle{plain}
\bibliography{newbib}

\end{document}